\def\year{2020}\relax
\documentclass[letterpaper]{article} 
\usepackage{aaai20}  
\usepackage{times}  
\usepackage{helvet} 
\usepackage{courier}  
\usepackage[hyphens]{url}  
\usepackage{graphicx} 
\usepackage{graphicx}  
\usepackage{multirow}
\usepackage{booktabs}
\usepackage{amsfonts}
\usepackage[colorlinks,linkcolor=red,anchorcolor=blue,citecolor=green]{hyperref}
\title{Deep Frequent Spatial Temporal Learning for Face Anti-Spoofing}
\author{Ying Huang\textsuperscript{\rm 1}, 
Wenwei Zhang\textsuperscript{\rm 1}, 
Jinzhuo Wang\textsuperscript{\rm 2}\thanks{Primarily Mike Hamilton of the Live Oak Press, LLC, with help from the AAAI Publications Committee}\\ 
\textsuperscript{\rm 1}Guangzhou Huya Information Technologies Co., Limited\\ 
\textsuperscript{\rm 2}Department of Engineering Science, University of Oxford\\ 
\{huanying, zhangwenwei1\}@huya.com, jinzhuo.wang@eng.ox.ac.uk 
}
 
\begin{document}

\maketitle

\begin{abstract}
Face anti-spoofing is crucial for the security of face recognition system, by avoiding invaded with presentation attack. 
Previous works have shown the effectiveness of using depth and temporal supervision for this task. 
However, depth supervision is often considered only in a single frame, and temporal supervision is explored by utilizing certain signals which is not robust to the change of scenes. 
In this work, motivated by two stream ConvNets, we propose a novel two stream FreqSaptialTemporalNet for face anti-spoofing which simultaneously takes advantage of frequent, spatial and temporal information. 
Compared with existing methods which mine spoofing cues in multi-frame RGB image, we make multi-frame spectrum image as one input stream for the discriminative deep neural network, encouraging the primary difference between live and fake video to be automatically unearthed. 
Extensive experiments show promising improvement results using the proposed architecture. 
Meanwhile, we proposed a concise method to obtain a large amount of spoofing training data by utilizing a frequent augmentation pipeline, which contributes detail visualization between live and fake images as well as data insufficiency issue when training large networks.
\end{abstract}

\noindent 
\section{Introduction}
 With the increasing number of applications making face image as unlock cues in our daily lives, the security issue of face spoofing has aroused great concern in the community. 
 Although face is an easily obtained and distinguished features for biometric system, it can also be easily mimicked.
 Presentation attack (PA), such as a simple printed photo (print attack), a digital replay video (replay attack) or wearing a mask (mask attack), could hack in a face recognition system if no face anti-spoofing module was equipped. 

\begin{figure}[t]
\centering
\includegraphics[width=0.9\columnwidth]{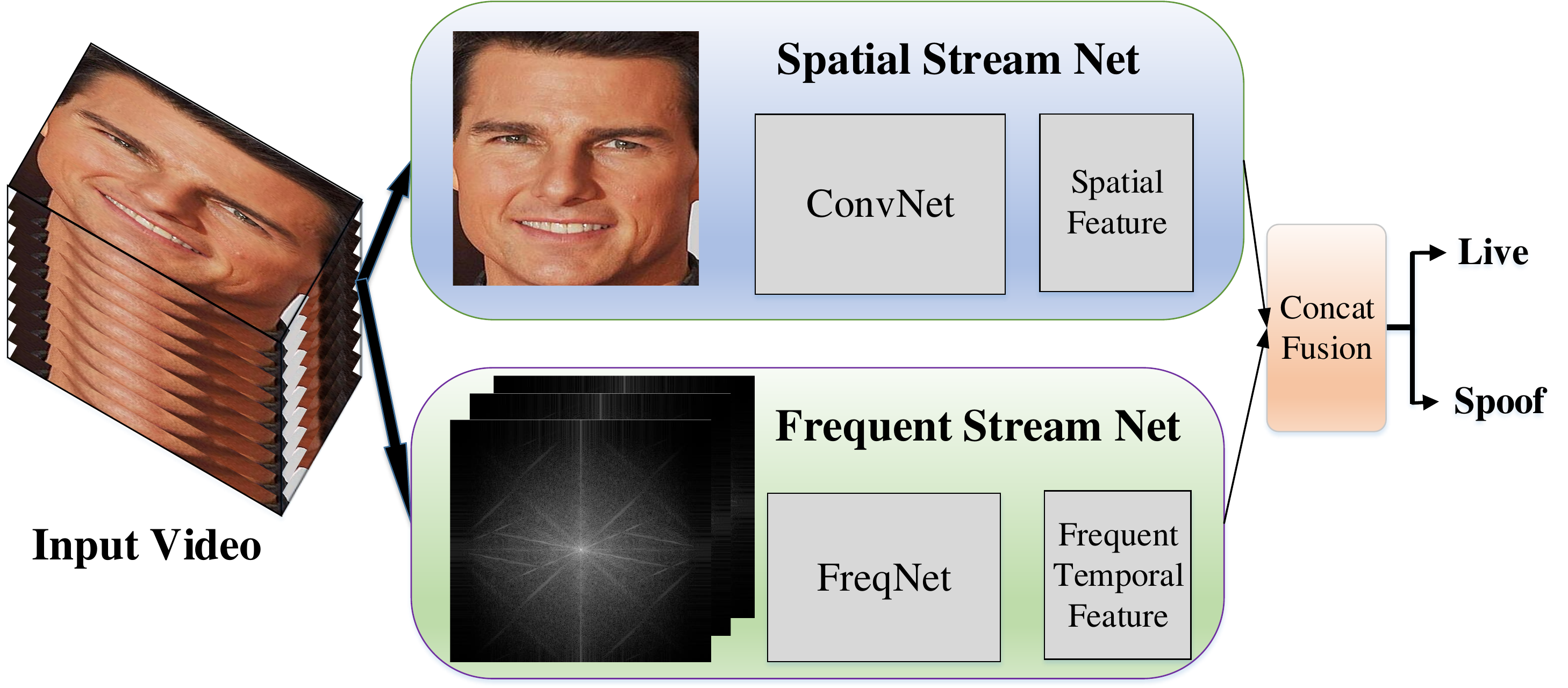} 
\caption{The pipeline for face anti-spoofing}
\label{fig_arch}
\end{figure}

To counteract presentation attack, a great number of methods \cite{li2004spie_fft,Erdogmus2014_3dmask,boulkenafet2016texture} have been proposed. In these methods, one or multi-frame RGB image is the standard input for the anti-spoofing system, with which handcrafted features \cite{maatta2011face,de2012lbp,tirunagari2015visual_dynamics,boulkenafet2016texture} or automatic learned deep features \cite{atoum2017depthPatch,tu2019learning} are extracted. 
Binary or auxiliary information \cite{liu2018Aux} is used to supervise the learning process. 
Due to the lack of an explicit correlation between pixels (color or texture) and attack patterns, extracting robust handcrafted features for different types of attack is challenging. On the other hand, deep learning methods have been widely used to explore uncertain correlations. However, in this task, it may fall in overfitting pitfall when little spoofing training data is presented \cite{yang2019faceData}. Instead of using only RGB frames as input, there are also efforts \cite{Hernandez2018nir-pulse,zhang2018casia,zhang2019feathernets} using near infrared spectroscopy and depth images as extra input source for anti-spoofing judgement. These methods can gain moderate improvement with the aid of additional input data due to the reduced uncertainty. Nevertheless, infrared and depth sensors are not default configuration in common mobile devices. Thus, these methods can not show their power in real deployment environment. 

Previous researches suggest that the main issue lies in the ubiquity of limit data source and the representational capacity of spoofing feature learning.
Is there a way that can utilize the ability of deep network and conquer the input lack of diversity? 
To solve this problem, as shown in Figure \ref{fig_arch}, we propose a two stream deep neural network architecture. By adding the frequent domain spectral image as extra input stream, the diversity of input can be increased. 
Depth image is used as label to supervise the deep network to learn robustly spatial features for different presentation attack patterns. 
Due to the intrinsic attribute of deep convolutional network, features in deep layer express high level semantic abstractions. 
This attribute goes against the goal of spoofing judgement when the fact is lied in low level pixels.
To overcome the shortage, the original low level information is reserved and passed to the classifier through the frequent part. 
On the other side, large scale training data is pivotal for a practical system.
However, spoof faces is hard to obtain in real environment and the new type of presentation attack emerge in endlessly. 
How to mimic these data quickly can affect the evolution speed of a defense system.
We proposed a technique to synthesize spoof faces through replacing blocks in their high frequent domain by blocks from spoof face. This technique can help us to solve the  shortage of data to a certain extent.
The main contributions of this work are summarized as follows:
\begin{itemize}
\item We propose a novel two stream architecture for end-to-end mining the primary difference between live and spoof video. 
This architecture improves the anti-spoof generalization ability by utilizing the spatial frequent temporal domain information simultaneously.
\item We propose to employ novel multi-frame spectral image  as network input, which let the original low level detail information easily pass to the final decision layer. And the experiment results show this setting can help to improve the distinguish ability of the learned deep model. 
\item We propose a concise technique to manufacture  presentation attack data,  which can help practical face anti-spoofing system to obtain large scale training data in a short time. And we demonstrate improved results and visualize the difference between live and spoof patterns.
\end{itemize}

\section{Related Work}

We review the previous face anti-spoofing works with their learning methods and discuss their relations.

{\bfseries Traditional Methods.}
Many prior works attempt to distinguish live and spoof faces from color and texture clues. 
A lot of hand-crafted features have been explored in these researches, including SLRB \cite{tan2010sparse}, LBP \cite{maatta2011face,de2012lbp}, HoG \cite{komulainen2013hog}, IDA \cite{wen2015distortion}, SIFT \cite{patel2016secure_unclock} and SURF \cite{boulkenafet2016surf}. And traditional classifiers such as Support Vector Machine (SVM) or Linear Discriminant Analysis (LDA) are adopted in these methods. Different data domains have been exploited to extract discriminative features, such as Fourier frequency spectrum domain\cite{li2004spie_fft} or HSV \cite{boulkenafet2016texture} color space.

Since one still image is weak on account of information gain, researchers attempt to leverage temporal change in face area. Optical flow \cite{bao2009liveness}, motion HOOF descriptor \cite{bharadwaj2013motion} and specular difference \cite{ebihara2019specular} are proposed as temporal-specific features for anti-spoofing judgement. Another line is to incorporate auxiliary information. For example, Remote photoplethysmography (rPPG) \cite{bobbia2016remotePPGskin,Liu2016rPPG-3dMask,nowara2017ppgsecure} is used as an message for 3d mask attack recognition. Compared to mask image sequence, the live faces exhibit a pulse of heart rate. So the heart rate signal can be estimated by enlarging the signal changes in the region of interest. Color or motion details are explored to extract rPPG signals. But rPPG signal becomes vulnerable with illumination changes of scene. Besides, these method may cost a long period to get an accurate rPPG sinal on the test time, which is unbearable for an app user.

{\bfseries Deep Learning Methods.}
Since AlexNet \cite{krizhevsky2012imagenet} was invented, the strong representation power of modern CNNs are proved in image recognition \cite{he2016resnet}, detection \cite{ren2015faster} and segmentation \cite{he2017mask} tasks. 
There are many recent works \cite{feng2016integration,li2016original,tu2019learning,song2019discriminative} using CNNs for face anti-spoofing. They regard face anti-spoofing as a binary classification problem and learn deep feature representation for single image. However, due to the simple binary supervision, these method are hard to generalize well in the cross-database testing. To overcome this weakness, patch and depth-based CNNs \cite{atoum2017depthPatch} utilize depth image as spatial supervision for deep features learning, which demonstrate the effectiveness of employing depth image as auxiliary label. Instead of using single depth map as network suppervision, \cite{wang2018temporal-depth} proposes to extract short-term and long-term motion by learning multi-frame temporal and depth information. AuxNet \cite{liu2018Aux} advances this method in a number of aspects, including fusion with temporal supervision by rPPG signal. \cite{shao2019multi} proposes to learn a generalized feature space via deep domain generalization framework. 
Overall, most existing methods tend to take one or multi-frame RGB image as CNN input and extract feature map from deep layer as feature representation for face anti-spoofing. Since in deep neural network, the feature maps from deep layer express high semantic level information, these method have a hard time to conquer the problem when fake cues lies in the low level image pixels, which is not unusual in real environment. 

\begin{figure*}[htb]
\centering
\includegraphics[width=0.9\textwidth]{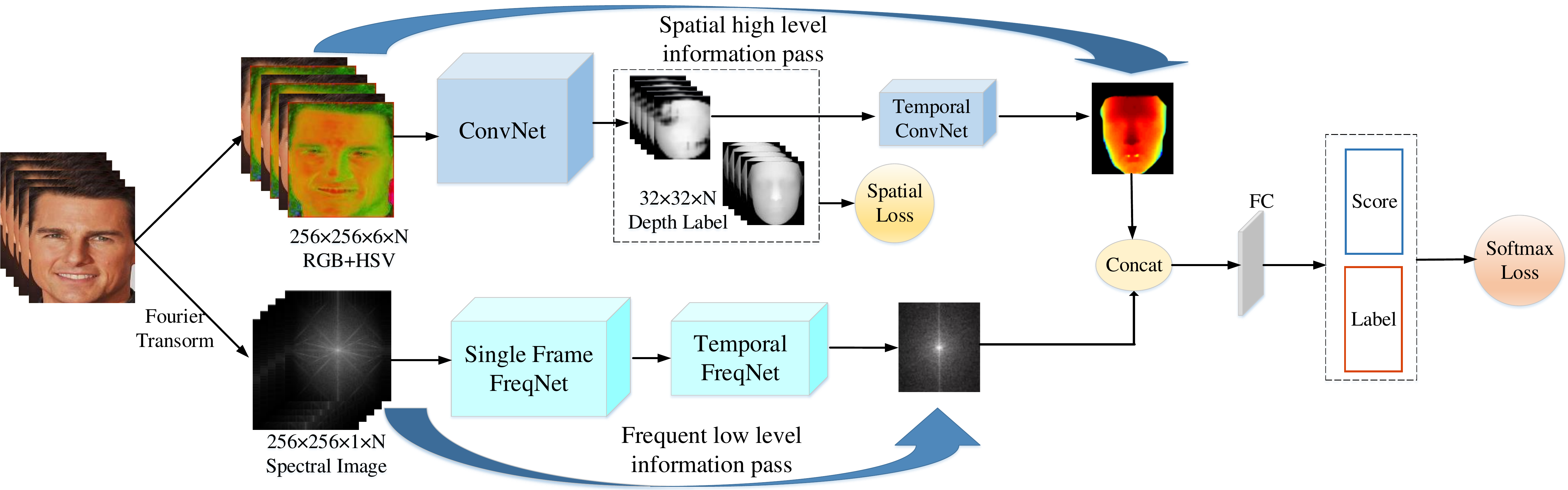} 
\caption{The proposed frequent spatial temporal architecture}
\label{fig_pipeline}
\end{figure*}

In this work, motivated by two-stream ConvNet \cite{Simonyan2014Two-stream-ConvNet} and two-step network \cite{wang2019video-inpainting}, we propose a novel two-stream architecture to learn the feature in different aspect and jointly train both sub-networks in an end-to-end manner. By employing multi-frame frequent spectrum images as supplement of the deep spatial part, the low level information can easily pass through the network to classifier, consequently overcoming the drawbacks in the aforementioned case and making this method generalize well in real-world testing. Another important issue is data insufficiency when training deep network. Previous researches tries to figure it out through different way. 
\cite{yang2019faceData} presents a data collection solution along with a data synthesis technique to simulate digital medium based face spoofing attacks.
Their experiment results show data synthesis solution is effective for deep network learning. But the synthesis pipeline is sophisticated and artificial, which may not mimic the practical data well in real environment.
Face de-Spoofing \cite{Liu2018FaceDeSpoofing} offers a perspective for detecting spoofing face from presentation attack by inversely decomposing a spoof face image into the live face and the spoofing noise, and a de-spoofing CNN architecture is proposed to estimate the spoofing noise.
Reverses the idea of face de-Spoofing, we propose to synthesize the negative presentation attack data by adding spoofing pattern to live face. In this way, attack image is obtained by randomly sampling from live and fake image pair in their high frequent domain, which make the synthesis more realistic and easy to implement.

\section{Proposed Method}
This section describes the framework and the details of our proposed method. 
Firstly, we introduce the substantial characteristics of the model. 
Then we elaborate the two stream network architecture we proposed and the learning strategy employed in the pipeline. 
Next we present the implementation details.
Finally, a data synthesis technique to obtain large scale training data is described.

\subsection{Model Characteristics}
The main idea of the proposed approach is to guide the deep network to focus on the obvious spoof patterns across different domain, automatically make a trade off in selecting the high level semantic deep features and low level frequent expression.
From previous work, the depth map supervision is proved powerful for deep convolutional network to learn spatial feature representation in face anti-spoofing task. 
Nevertheless, there are cases which always fail in spatial domain. 
So we can not improve the model efficiency by only reinforcing this information. 

Temporal changes of input source can be utilized at this moment. 
We employed it by taking multi-frame images as network input.
Due to the different level of semantic mapping in disparate layer of deep network, a shortcut is needed to transfer the original input information to the classifier when we hope the classifier to mine the spoofing cues in low level pixels. 
Fourier transform is employed to get the spectral image, which keep the whole input information but expressed it in a different way.

Let us consider $F(x)$ as an underlying mapping between input source and the feature representation. 
And $H(x)$ is the measurement of information in $x$.
Then the CNN model would learn a mapping $F_{spatial}$ between the input face and the spatial representation layer.
The hidden semantic information, there is different spatial depth lies in live and fake faces, is expressed by this mapping.
On the other side, the frequent mapping $F_{frequent}$ retains the original input information. This can be expressed as the formulation:
\begin{displaymath}
    F_{frequent}(x) = y, H(x) = H(y)
\end{displaymath}

Since the original input information is presented in an easily distinguished view for the classifier, that make the classifier to be simple and reduce the risk of overfitting.

\subsection{Network Architecture And Training Strategy}

The proposed architecture take advantage of two stream input as shown in Figure \ref{fig_pipeline}. 
In the process of network learning, the top ConvNet stream evaluates each frame separately and estimates the depth maps for them. 
And a sub Temporal ConvNet learns on this sequence of depth maps to generate the feature representation in spatial temporal domain.
At the same time, the bottom Temporal FreqNet collects the spectrum images and evaluates the temporal difference across the input of sequence automatically. 
A Frequent Temporal feature representation in low semantic level is generated through a shallow network on these frames.
Finally, the separated spatial part and temporal frequent part are concatenated together to form the final feature representation. 
And a fully connected layer convert the representation from feature space to category space. A simple classifier learns on this feature to judge whether a face is live or spoof.

We adopt the general four block 18-layer ResNet as the backbone for the ConvNet.
While this backbone can be replaced by other convolutional network, such as VGG, Densenet, EfficientNet, MobileNet and so on.
The detail of the whole network architecture is described in Table \ref{model_param}.
\begin{table}[ht]
\caption{The network structure of Spatial ConvNet, FreqTempNet and TemporalConvNet.}\smallskip
\centering
\resizebox{.95\columnwidth}{!}{
\smallskip\begin{tabular}{ccc||ccc}
\hline \hline
    Layer & Filter/Stride/kernel & Output Size & Layer & Filter/Stride & Output Size  \\ \hline \hline
     & \multirow{2}*{SpatialConvNet} &  &  & \multirow{2}*{FreqTempNet} &  \\ 
     & & & & & \\ \hline
     & rgb+hsv & 256 & & spectrogram  & 256 \\ \hline 
    Conv0-0 & 64/2 & 128 & InstanceNorm & & 256      \\ 
    MaxPool1  &    & 64  & DepthWiseConv& 5/1 & 256             \\ 
    Conv1-1 & 64/1 & 64  & MaxPool2 & -/8 & 32  \\ 
    Conv1-2 & 64/1 & 64  & PointWiseConv & 1/1  & 32 \\  
    Conv1-3 & 64/1 & 64  & LayerNorm & & 32 \\ 
    Conv1-4 & 64/1 & 64  & MaxPool3 & -/2 & 16 \\ \hline

    Conv2-1 & 128/1 & 64 & &  & \\ 
    Conv2-2 & 128/1 & 64 & &  &  \\ \cline{4-6}
    Conv2-3 & 128/1 & 64 & & \multirow{2}*{TemporalConvNet} & \\
    Conv2-4 & 128/2 & 64 & &  & \\ \hline

    Conv3-1 & 256/1 & 32 & Conv-0 & 1/1 & 32   \\ 
    Conv3-2 & 256/1 & 32 & MaxPool4 & -/2 & 16 \\ \cline{4-6}
    Conv3-3 & 256/1 & 32 & & \multirow{2}*{FeatureLayer} &  \\ 
    Conv3-4 & 256/1 & 32 & & & \\ \hline

    Conv4-1 & 512/1 & 32 & MultiSpatialMap &  & 16 \\ 
    Conv4-2 & 512/1 & 32 & MultiFreqMap & & 16 \\ 
    Conv4-3 & 512/1 & 32 & Concat & &  \\ 
    Conv4-4 & 512/1 & 32 & FC & & 512 \\ \hline \hline
\end{tabular}
}
\label{model_param}
\end{table}
Given a sequence of $N$ frame color image $I^{256\times256\times3}$ as input, we first compute their frequent information via Fourier transform and get their spectral image $I_f^{256\times256}$ . 
Then fed this sequence of consecutive $N$ frame $I$ to the network.
Followed the setting in \cite{Liu2018FaceDeSpoofing}, we use rgb+hsv as the spatial ConvNet stream input and their corresponding $N_f$ spectrum images as the input of FreqTempNet stream.

There are two loss employed in the learning process.
Depth loss $\Theta_{depth}$ for spatial supervision and the softmax loss $\Theta_c$ for classifying error. 
PRNet \cite{Feng2018PRNet} is used to generate depth label for live faces and we normalize them in a range of $[0,1]$.
While setting spoofing depth map to 0.
The estimated depth map is resized to a pre-defined size of $32\times32$, make $Label_{depth}\in\mathbb{R}^{32\times32}$.
\begin{displaymath}
    \Theta_{depth}=\Arrowvert F_{spatial}(x) - Label_{depth}\Arrowvert
\end{displaymath}

As mentioned before, this two stream architecture was designed to make frequent part to be supplementary of the spatial stream. 
Trying to conquer the failure when spoof pattern lies in low level image pixels. 
To achieve this goal, we utilize a training strategy from \cite{he2019bag} to harmonize the two parts. 
We add a learning rate warmup period for the parameters in frequent temporal part, setting the learning rate of this part start from zero and gradually increases it to the target value.
On the other side, the learning rate for the backbone Spatial-ConvNet is setting to a large value and gradually decreases to zero.
This configuration make the network firstly to focus on the primary spatial semantic information for face anti-spoofing and try it best to give its judgement. 
At the later learning stage, the parameters for Spatial-ConvNet is fixed, lead to the spatial representation unchanged. 
And low level frequent  stream reveal its power to overcome the fail case in depth estimating.
Then the whole framework make a trade off in selecting between high level semantic information and low level color texture detail automatically.
While there are several parts in the whole network architecture, it can be trained in an end-to-end manner.

\subsection{Data Synthesis Pipeline}
Data plays an important role for a robust model learning. 
Typically the training data is expected to be in a large scale and close to the testing data. 

However, current database in the community are either of small scale or far from real world testing data. 
So a model can not generalize well with these limited data. 
On the other hand, presentation attack is an abnormal behavior. 
Compared to the positive live faces can be easily downloaded from internet, the negative fake data may cost many time to gain. 
In \cite{yang2019faceData}, they proposed a method to mimic the negative examples through zoom, add reflection and perspective transformation on live faces. 
And their experiment results show that the mimic data is beneficial for the ability of model generalization. 
But their mimic pipeline is sophisticated. 
And these manually defined step may break the data distribution in the real.
Is there a brief method to implement these process while keep the data distribution? 

\begin{figure}[ht]\centering
\includegraphics[width=0.9\columnwidth]{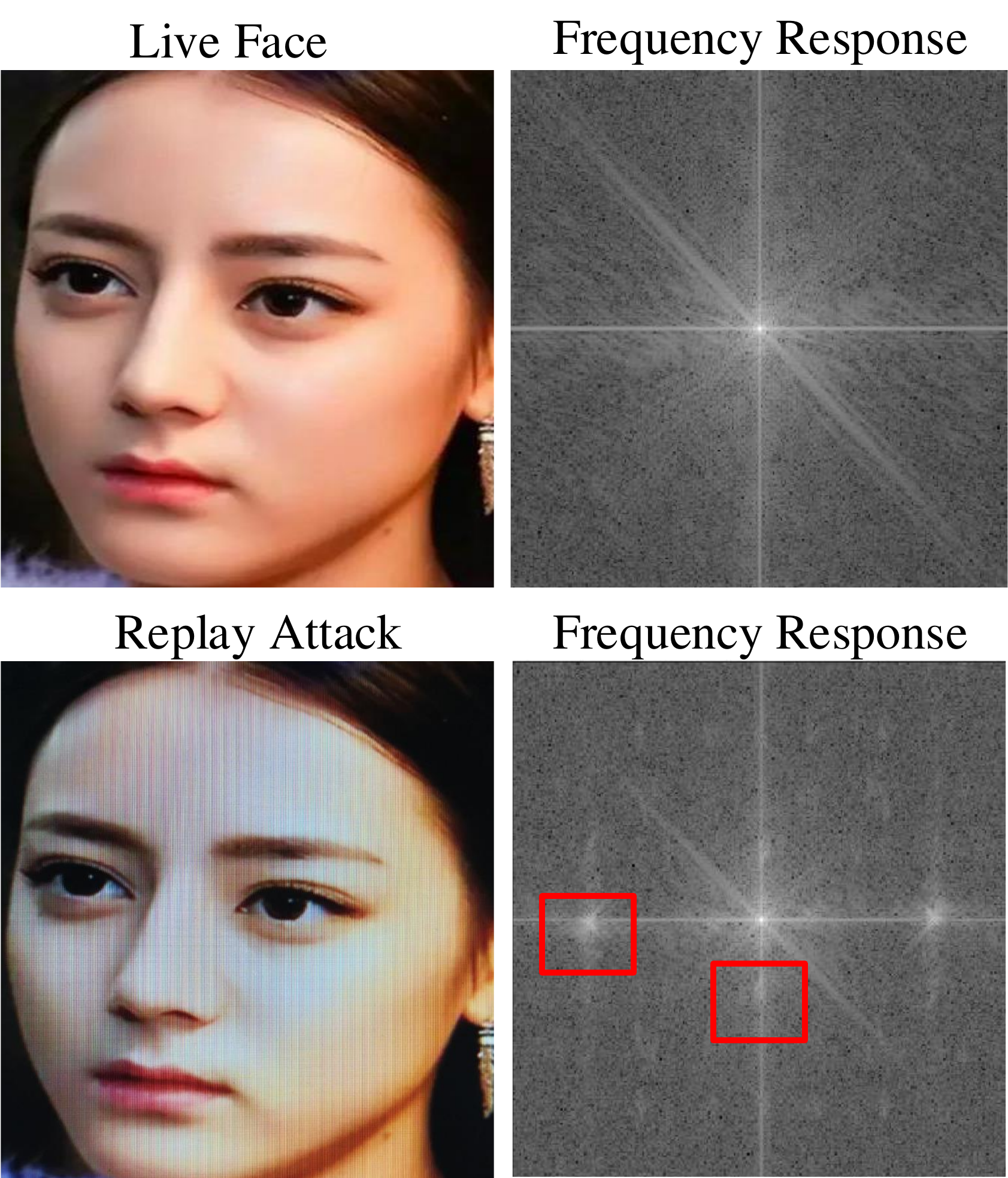} 
\caption{The spoof pattern between live and fake face. {\bfseries Top:} live face and its frequency response. {\bfseries Bottom:} spoof face, we can visualize the spoof pattern hidden in high frequency domain.}
\label{fig_spoof}
\end{figure}
With a set of manually mimicked spoof faces in hand, we can observe the spoof pattern behind them. 
Human judge whether a face is spoofing by carefully check the detail of the image. 
There will be color distortion, display artifacts or presenting noise in fake image, which were expressed in high frequency domain of their spectral images.
We show an example in Figure \ref{fig_spoof}. 
There are moire pattern and color distortion in the spoof face.
These effects are hard to simulate by a pre-define step in spatial domain. 
But we can obtain these spoof patterns by sampling blocks in spectral image. 
If we replace these area in spectral image of live face with the sampled ones, then the spoof pattern is transferred to live face. 
This process is very concise and keep the data distribution unchanged.

\begin{figure*}[t]
\centering
\includegraphics[width=0.8\textwidth]{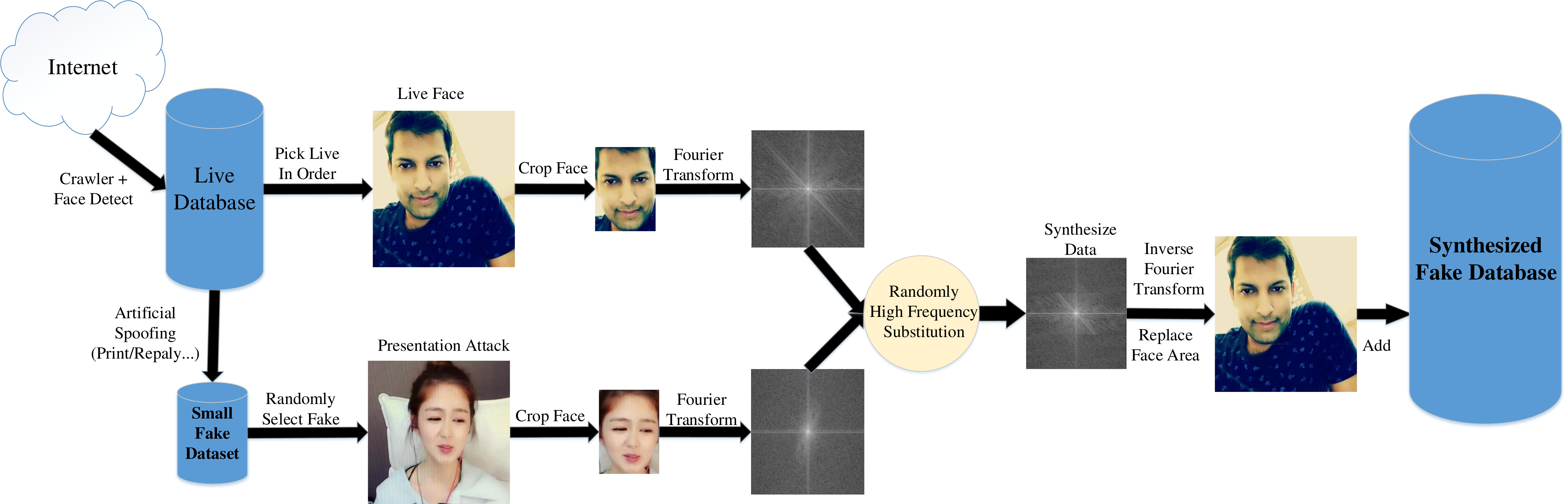} 
\caption{The data synthesis pipeline}
\label{fig_data_aug}
\end{figure*}

Follow this way, we propose a pipeline to collect large scale database. 
See the details in Figure \ref{fig_data_aug}. 
We firstly collected a large scale live face database of ten thousand people by crawling video from internet and filtered them by a face detector.

Then a small percent of these face videos were selected and a spoof dataset was manually simulated by carefully handling on these videos.
We used various kind of digital camera (including iPhone, Huawei P10, Huawei Honer V10, Xiaomi 4 and Redmi Note) to record the video to obtain the replay attack videos. 
At the same time, printed one hundred face image and recorded them by previous different video camera to get print attack videos. 
Then the concise simulate step is repeated on each single face image by randomly replacing some blocks of their spectrum image with the corresponding blocks from spoofing one. 
Note that for a sequence of image in one video, the areas needed to be replaced were randomly selected in the first frame. 
And the same area in the following frames is replaced by the corresponding area of the following fake images. 
This setting leave the data distribution in temporal domain unchanged.
Through this method, we can obtain large scale fake database for any type of presentation attack with little cost. 
When a new spoof type appeared in real test environment, we can produce a large scale database for that type with different person id in a short time. 
This is very helpful for the evolution of modern face anti-spoofing system.

\section{Experimental Results}
In this section, we conduct extensive experiments on different datasets. The experimental setup, ablation study, testing results and analysis are illustrated in sequence.

\subsection{Experimental Setup}
We first describe the datasets, evaluation metrics and implementation details for all the experiments.

{\bfseries Datasets.}
The proposed model is evaluated on three public face anti-spoofing databases, including OULU-NPU \cite{boulkenafet2017oulu}, CASIA-SURF \cite{zhang2018casia} and SiW \cite{liu2018Aux}. 
We use SiW and OULU-NPU for cross testing and comparing with the state of art intra testing result on these datasets.

{\bfseries Evaluation metrics.}
The following evaluation metrics: Attack Presentation Classification Error Rate (APCER), Bona Fide Presentation Classification Error Rate (BPCER) and Half Total Error Rate (HTER) are employed to compare with prior works. 
The HTER is half of the sum for the False Rejection Rate (FRR) and the False Acceptance Rate (FAR).
Besides, ACER = (APCER+BPCER)/2 is used to report our results.
To compare the model distinguish ability under different threshold, we report the results of True Positive Rate (TPR) at different False Positive Rate (FPR).

{\bfseries Implementation details.}
The proposed method is implemented in MXNet \cite{chen2015mxnet}. 
And the four block ResNet-v1b-18 \cite{he2019bag} is adopted as backbone networks for spatial part. 
SpatialConvNet models are trained by learning rate 0.3 with a cosine learning rate decay and weight decay 1e-4 by the supervision of the depth loss. 
The FreqTempNet stream is trained with learning rate 0.03 and weight decay 1e-5. 
All models iterate 10 epochs in the training stage while 5 epoch is setting as warmup period for the FreqTempNet . 
The batch size of both spatial and frequent stream is 16 with sequence length 10 as general setting. 
We random initialize the whole network by using as normal distribution with zero mean and std of 0.02.

\subsection{Ablation Study}

\begin{table}[t]
\caption{Different model results on OULU Protocol 2}\smallskip
\centering
\resizebox{.95\columnwidth}{!}{
\smallskip\begin{tabular}{ccccc}
\toprule
    Method & APCER & BPCER & ACER  \\ \hline \hline
    Depth Model &  $10.2$ & $7.8$ & $14.0$ \\ 
    FreqSpatialNet & $4.9$ & $3.3$ & $4.1$ \\ 
    FreqSpatialTempNet1(5 Frames)  & $2.9$ &  $1.8$ &  $2.4$ \\ \hline
    FreqSpatialTempNet2(10 Frames) & $\mathbf{2.3}$ & $1.0$ & $\mathbf{1.7}$ \\ 
    FreqSpatialTempNet3(20 Frames) & $2.8$ & $\mathbf{0.8}$ & $1.8$ \\ \bottomrule
\end{tabular}
}
\label{model_part}
\end{table}

{\bfseries Advantage of proposed architecture}
We compare three architectures to demonstrate the advantages of the proposed framework. 
And four different choices of sequence length is compared to show the advantage of using longer input image. 
The experiment results is shown in Table \ref{model_part}.
The \textit{Depth Model} use only spatial stream to learning deep convolutional features, without the low level frequent information in the temporal frequent stream. 
This is implemented by setting the spectrum image to zeros.
\textit{FreqSpatialNet} has a similar architecture to the proposed method, except the input image number is set as 1. 
So it can not explore the clues on temporal domain. 
\textit{FreqSpatialTempNet1} is the proposed architecture with setting the input length of sequence as 5.
While change the input frame length to 10 and 20, we get \textit{FreqSpatialTempNet2} and \textit{FreqSpatialTempNet3}. 
This help us to compare the effectiveness of different temporal length. We train all the model with the OULU-NPU training set under protocol 2 and the APCER, BPCER and ACER values are reported in testing set. 
As shown in Table \ref{model_part}, the FreqSpatialNet result is $4.1\%$ (ACER) compared to the $14.0\%$ (ACER) of Depth Model. 
This demonstrate the effectiveness of the feature combine mode. 
With setting the input image length to 10 and 20, we achieve $1.7\%$ (ACER) in FreqSpatialTempNet2 and $1.8\%$ (ACER) in FreqSpatialTempNet3. 
They exceed the result of FreqSpatialTempNet1. 
This shows the network can take advantage of longer sequence input.

\begin{table}[t]
\caption{Performance in CASIA-SURF validation dataset}\smallskip
\centering
\resizebox{.95\columnwidth}{!}{
\smallskip\begin{tabular}{cccc}
\hline
Method & ACER & TPR@FPR=10E-2 & TPR@FPR=10E-3  \\ \hline
ResNet18-Baseline   & $0.05$    & $0.883$  & $0.272$ \\
FeatherNet & $\mathbf{0.00261}$ & $\mathbf{1.0}$    & $\mathbf{0.9615}$ \\ 
Ours & $0.016$ & $0.958$ & $0.763$ \\ \hline
\end{tabular}
}
\label{casia_tabel}
\end{table}

\begin{table}[t]
\caption{Results of intra-testing on OULU Protocols}\smallskip
\centering
\resizebox{.95\columnwidth}{!}{
\smallskip\begin{tabular}{|c|c|c|c|c|}
\hline
Prot. & Method & APCER(\%) & BPCER(\%) & ACER(\%) \\ \hline \hline
\multirow{3}*{1} & AuxNet &  $\mathbf{1.6}$  & $1.6$ & $1.6$ \\ \cline{2-5}
~ & Temporal-Depth          & $2.5$ & $0.0$ & $1.3$ \\ \cline{2-5}
~ & Ours                    & $2.2$ & $\mathbf{0.0}$ & $\mathbf{1.1}$ \\ \hline \hline

\multirow{3}*{2} & AuxNet & $2.7$ & $2.7$ & $2.7$ \\ \cline{2-5}
~ & Temporal-Depth & $\mathbf{1.7}$ & $2.0$ & $1.9$ \\ \cline{2-5}
~ & Ours & $2.3$ & $\mathbf{1.0}$ & $\mathbf{1.7}$ \\ \hline \hline

\multirow{3}*{3} & AuxNet & $\mathbf{2.7\pm1.3}$ & $3.1\pm1.7$ & $\mathbf{2.9\pm1.5}$ \\ \cline{2-5}
~ & Temporal-Depth      & $5.9\pm1.9$ & $5.9\pm3.0$ & $5.9\pm1.0$ \\ \cline{2-5}
~ & Ours                & $4.4\pm3.7$ & $\mathbf{2.0\pm1.4}$ & $3.2\pm1.4$ \\ \hline \hline

\multirow{3}*{4} & AuxNet & $9.3\pm5.6$ & $10.4\pm6.0$ & $9.5\pm6.0$ \\ \cline{2-5}
~ & Temporal-Depth & $14.2\pm8.7$ & $\mathbf{4.2\pm3.8}$ & $9.2\pm3.4$ \\ \cline{2-5}
~ & Ours            & $\mathbf{9.3\pm3.7}$ & $7.5\pm3.7$ & $\mathbf{8.4\pm2.7}$ \\ \hline
\end{tabular}
}
\label{oulu_tabel}
\end{table}

\subsection{Intra Testing}
We compare the performance of testing result on CASIA-SUFR, OULU-NPU and SiW datasets. 
We follow the four protocols on OULU-NPU and three protocols on SiW for the testing. 
The metric value of APCER, BPCER, TPR at different FRR and ACER are reported to compare the proposed model and other methods on these datasets. 

In Table \ref{casia_tabel}, we train our model with CASIA-SURF training set and tested on its validation set, compared with ResNet18-Baseline \cite{zhang2018casia} and FeatherNet \cite{zhang2019feathernets} results. 
Though the results is not as good as the FeathreNet, the experiment result is meaningful. Since both the ResNet18-Baseline and FeatherNet used RGB, IR and Depth image as network input, however our method used only RGB images. As mentioned before, this is more common hardware setting for mobile devices.

Table \ref{oulu_tabel} shows our method achieves the lowest ACER in 3 out of 4 protocols on OULU-NPU dataset. 
The ACER result is slightly worse than that of AuxNet on protocol 2. 
And the intra testing experiments result on three protocols of SiW in showed in Table \ref{siw_tabel}. We achieve lowest ACER in the protocol 1 and lowest APCER in protocol 2.

\begin{table}[t]
\caption{Results of intra-testing on three protocols of SiW}\smallskip
\centering
\resizebox{.95\columnwidth}{!}{
\smallskip\begin{tabular}{|c|c|c|c|c|}
\hline
Prot. & Method & APCER(\%) & BPCER(\%) & ACER(\%) \\ \hline \hline
\multirow{3}*{1} & AuxNet & $3.58$ & $3.58$ & $3.58$ \\ \cline{2-5}
~ & Temporal-Depth & $0.96$ & $\mathbf{0.50}$ & $0.73$ \\ \cline{2-5}
~ & Ours & $\mathbf{0.8}$ & $\mathbf{0.50}$ &  $\mathbf{0.67}$\\ \hline \hline

\multirow{3}*{2} & AuxNet & $0.57\pm0.69$ & $0.57\pm0.69$ & $0.57\pm0.69$ \\ \cline{2-5}
~ & Temporal-Depth & $0.08\pm0.17$ & $\mathbf{0.21\pm0.16}$ & $\mathbf{0.15\pm0.14}$ \\ \cline{2-5}
~ & Ours        & $\mathbf{0.00\pm0.00}$ & $0.75\pm0.96$ &  $0.38\pm0.48$\\ \hline \hline

\multirow{3}*{3} & AuxNet & $8.31\pm3.81$ & $8.31\pm3.80$ & $8.31\pm3.81$ \\ \cline{2-5}
~ & Temporal-Depth & $\mathbf{3.10\pm0.79}$ & $\mathbf{3.09\pm0.83}$ & $\mathbf{3.10\pm0.81}$ \\ \cline{2-5}
~ & Ours & $9.5\pm1.2$ & $5.3\pm2.1$ & $7.4\pm2.9$\\ \hline
\end{tabular}
}
\label{siw_tabel}
\end{table}

\begin{table}[t]
\caption{Results of cross-testing on SiW and OULU}\smallskip
\centering
\resizebox{.95\columnwidth}{!}{
\smallskip\begin{tabular}{|c|c|c|c|c|}
\hline
Train & Test & Method &  ACER(\%) \\ \hline \hline
\multirow{8}*{SiW} & \multirow{2}*{Oulu1} & AuxNet & $10.0$ \\ \cline{3-4}
~ & ~ & Ours & $\mathbf{9.3}$ \\ \cline{2-4}

~ & \multirow{2}*{Oulu2} & AuxNet & $14.1$ \\ \cline{3-4}
~ & ~ & Ours & $\mathbf{7.8}$ \\ \cline{2-4} 

~ & \multirow{2}*{Oulu3} & AuxNet & $\mathbf{13.8\pm5.7}$ \\ \cline{3-4}
~ & ~ & Ours & $16.2\pm5$ \\ \cline{2-4}

~ & \multirow{2}*{Oulu4} & AuxNet & $\mathbf{10.0\pm8.8}$ \\ \cline{3-4}
~ & ~ & Ours & $14.1\pm8.3$ \\ \hline \hline

\multirow{2}*{SiW1} & \multirow{4}*{SiW1} & AuxNet & $3.58$  \\ \cline{3-4}
~                   & ~    &  Temporal-Depth & $0.73$ \\ \cline{1-1} \cline{3-4}
OULU                & ~    & Ours            & $7.28$ \\ \cline{1-1} \cline{3-4}
OULU+SynthesizedData & ~   & Ours+          & $\mathbf{0.71}$ \\ \hline \hline 

\multirow{2}*{SiW2} & \multirow{4}*{SiW2} & AuxNet & $0.57\pm0.69$  \\ \cline{3-4}
~ &  ~ & Temporal-Depth & $\mathbf{0.15\pm0.14}$ \\ \cline{1-1} \cline{3-4}
OULU                & ~  & Ours          & $6.9\pm1.1$ \\ \cline{1-1} \cline{3-4}
OULU+SynthesizedData & ~ & Ours+        & $0.53\pm0.26$ \\ \hline \hline

\multirow{2}*{SiW3} & \multirow{4}*{SiW3} & AuxNet & $8.31\pm3.81$ \\ \cline{3-4}
~ &     ~ & Temporal-Depth & $\mathbf{3.10\pm0.81}$ \\ \cline{1-1} \cline{3-4}
OULU & ~ &  Ours             & $11.6\pm4.7$ \\ \cline{1-1} \cline{3-4}
OULU+SynthesizedData & ~ & Ours+ & $8.02\pm1.63$ \\ \hline 
\end{tabular}
}
\label{cross_tabel}
\end{table}

\subsection{Cross Testing}
While intra-testing can show the effectiveness of the proposed method, cross testing is a touchstone for model generalization. 
In cross testing, the training and testing set is selected from different database. 
This setting will eliminate the bias from the limitation for data collection.
In the testing period, the person id and  spoofing sensor are guaranteed differ from the training set under cross testing.
And it is more similar to the situation a face anti-spoofing system will meet in real environment.

We reported the cross-testing results on SiW and Oulu-NPU dataset with different training dataset setting and compared the performance with the intra-testing results from two other methods in Table \ref{cross_tabel}.
We see our method achieve lower ACER in protocol 1 and 2 in OULU-NPU testing set compared with the results from AuxNet. 
This result shows the proposed method can grasp the different data distribution from live and and spoofing faces. 
Compared with the intra testing result On 3 protocols of SiW from AuxNet and Temporal-Depth, our method is just slightly worse than them, while we using cross-testing configuration.
This is worthy of attention, since cross testing is known to be essentially harder than intra testing. 

We generated a synthesized dataset by randomly selecting 80 spoof videos from SiW training set as source data. And handle them through the synthesize technique we introduced before in section data synthesis pipeline. 
Then the proposed method is trained with OULU-NPU plus this synthesized dataset and testing it on three protocol of SiW, the experiment results is showed in Table \ref{cross_tabel} as \textit{OUR+}. 
From the result, we can see that these synthesized data help our method achieve a performance that can compete with the results from intra-testing. 
This result shows that the proposed data synthesized technique can transfer the spoofing cues to live videos. 
And we can employ it to obtain large scale spoof video for practical face anti-spoofing systems.

\begin{figure}[t]
\centering
\includegraphics[width=0.9\columnwidth]{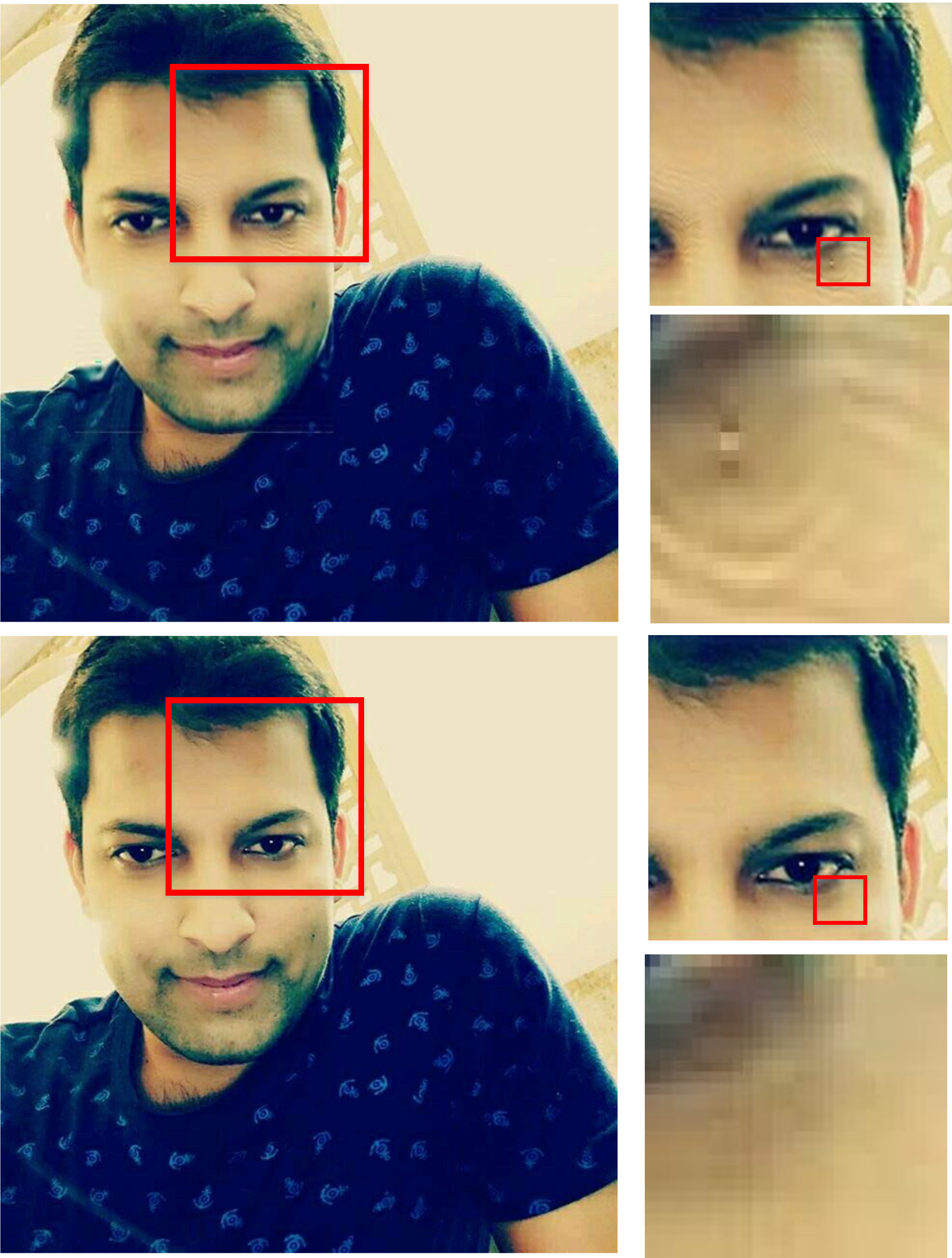} 
\caption{The detail between live and synthesized fake image. {\bfseries Top:} synthesized image and its local regions. {\bfseries Bottom:} live face crawled from internet. Moire pattern appears in the synthesized image. Best viewed electronically.}
\label{fig_detail}
\end{figure}

\subsection{Visualization and Analysis}
In the proposed architecture, the original spectrum images is used to overcome the shortcoming in spatial domain. 
And we proposed to replace the blocks from live face by the corresponding block from fake image. 
We are curious how will the synthesized image look like in spatial domain?
Figure \ref{fig_detail} shows an example of the live and synthesis pair.
The top is the synthesized image with the face area generated by the proposed technique. 
The bottom is the original live face and its local regions.
By comparing the details between synthesized fake image and live one, 
we can find fake clues appeared in generated face local region while the original person appearance left unchanged. 
Since the proposed architecture will learn the detail difference between them via two domain of input. 
The fake clues from the spoofing video are passed to the classifier through both depth estimation and frequent stream.

\section{Conclusions}
This paper proposed a practical architecture to build a robust face anti-spoofing system. 
The model, namely FreqSpatialTempNet, utilizes both the spatial, temporal and frequent domain information simultaneously. 
Mining the robust high level deep features and low level frequent features to distinguish live and spoof faces.
In addition, a concise data synthesized technique is presented to obtain large scale training data. 
It can help us break through the dilemma to some extent in which new presentation attack type emerge in endlessly and training data shortage is everywhere.
By conducting extensive experiments on three public face anti-spoofing datasets, the performances of the proposed method demonstrate better generalization ability compared with other most existing methods.

\bibliography{paper}
\bibliographystyle{aaai}

\end{document}